\documentclass[11pt,a4paper]{article}
\usepackage[hyperref]{acl2021}
\usepackage{times}
\usepackage{latexsym}
\usepackage{tabularx}
\usepackage{multirow}
\usepackage{graphicx}
\usepackage{booktabs}
\usepackage{subfig}
\usepackage{amsthm}
\usepackage{graphicx}
\usepackage{float}

\usepackage{microtype}

\aclfinalcopy % Uncomment this line for the final submission
\title{CUGE: A Chinese Language Understanding and \\ Generation Evaluation Benchmark}

\author{Yuan Yao$^{1}$, Qingxiu Dong$^{2}$, Jian Guan$^{1}$, Boxi Cao$^{3}$, Zhengyan Zhang$^{1}$, Chaojun Xiao$^{1}$\\

\textbf{Xiaozhi Wang$^{1}$, Fanchao Qi$^{1}$, Junwei Bao$^{4}$, Jinran Nie$^{5}$, Zheni Zeng$^{1}$, Yuxian Gu$^{1}$} \\

\textbf{Kun Zhou$^{6}$, Xuancheng Huang$^{1}$, Wenhao Li$^{1}$, Shuhuai Ren$^{2}$, Jinliang Lu$^{7}$, Chengqiang Xu$^{1}$} \\ 

\textbf{Huadong Wang$^{1}$, Guoyang Zeng$^{1}$, Zile Zhou$^{8}$, Jiajun Zhang$^{7}$, Juanzi Li$^{1}$, Minlie Huang$^{1}$} \\

\textbf{Rui Yan$^{8}$, Xiaodong He$^{4}$, Xiaojun Wan$^{9}$, Xin Zhao$^{6}$, Xu Sun$^{2}$, Yang Liu$^{1}$} \\

\textbf{Zhiyuan Liu$^{1*}$, Xianpei Han$^{3*}$, Erhong Yang$^{5*}$, Zhifang Sui$^{2*}$, Maosong Sun$^{1*}$}\\

\scalebox{0.92}{$^{1}$Department of Computer Science and Technology, Tsinghua University} \\
\scalebox{0.92}{$^{2}$MOE Key Lab of Computational Linguistics, School of EECS, Peking University} \\
\scalebox{0.92}{$^{3}$Institute of Software, Chinese Academy of Sciences \ $^{4}$JD AI Research, Beijing, China} \\
\scalebox{0.92}{$^{5}$School of Information Science, Beijing Language and Culture University} \\
\scalebox{0.92}{$^{6}$School of Information, Renmin University of China} \\
\scalebox{0.92}{$^{7}$National Laboratory of Pattern Recognition, Institute of Automation, CAS}\\
\scalebox{0.92}{$^{8}$Gaoling School of Artificial Intelligence, Renmin University of China} \\
\scalebox{0.92}{$^{9}$Wangxuan Institute of Computer Technology, Peking University} \\
\scalebox{0.92}{Beijing Academy of Artificial Intelligence}
}

\date{}

\begin{document}
\maketitle
\begin{abstract}
Realizing general-purpose language intelligence has been a longstanding goal for natural language processing, where standard evaluation benchmarks play a fundamental and guiding role. 
% Existing standard evaluation benchmarks typically evaluate models on a collection of datasets and average model performance across datasets as the overall score. 
We argue that for general-purpose language intelligence evaluation, the benchmark itself needs to be comprehensive and systematic. To this end, we propose \textbf{CUGE}, a \textbf{C}hinese Language \textbf{U}nderstanding and \textbf{G}eneration \textbf{E}valuation benchmark with the following features: (1) Hierarchical benchmark framework, where datasets are principally selected and organized with a language capability-task-dataset hierarchy. (2) Multi-level scoring strategy, where different levels of model performance are provided based on the hierarchical framework. To facilitate CUGE, we provide a public leaderboard that can be customized to support flexible model judging criteria. Evaluation results on representative pre-trained language models indicate ample room for improvement towards general-purpose language intelligence. CUGE is publicly available at \url{cuge.baai.ac.cn}.

\end{abstract}

{\let\thefootnote\relax\footnotetext{$*$ Corresponding Authors: Z. Liu (liuzy@tsinghua. edu.cn), X. Han (xianpei@iscas.ac.cn), E. Yang (yerhong@blcu.edu.cn), Z. Sui (szf@pku.edu.cn), M. Sun (sms@tsinghua.edu.cn)}}

{\let\thefootnote\relax\footnotetext{1. Update Note (April 14th, 2022): We add two new datasets, including grammatical error correction dataset YACLC from Beijing Language and Culture University, and reading comprehension dataset GCRC from Shanxi University, and also improve the description consistency of all datasets.}}

\section{Introduction}
% 介绍CUGE背景: 通用语言智能
% - 通用语言智能应该在多个能力和任务具有有效性，而非针对任务、数据集定制
% - 通用语言智能的发展需要一个科学系统的的标准benchmark来评测指导。尤其是对于预训练语言模型。
% 相关工作：GLUE/SuperGLUE/CLUE
% 不足：
% - 评测体系：并不全面系统；扁平结构，以数据集为中心
% - 评测方案：简单平均。没有多层次的能力-任务-数据集得分；没有考虑到数据、指标差异
% - 数据集评价：没有principle的评价方法
% 我们提出GUGE：
% - 全面系统的评测体系
% - 多层次维度的评测方案
% - 更加科学可靠的数据集评价方法：难度、信度、效度
% 简要概括设计原则、框架、评测方案、网站等
% 简要介绍实验
% 贡献总结

\begin{figure}[t]
    \centering
    \includegraphics[width=\columnwidth]{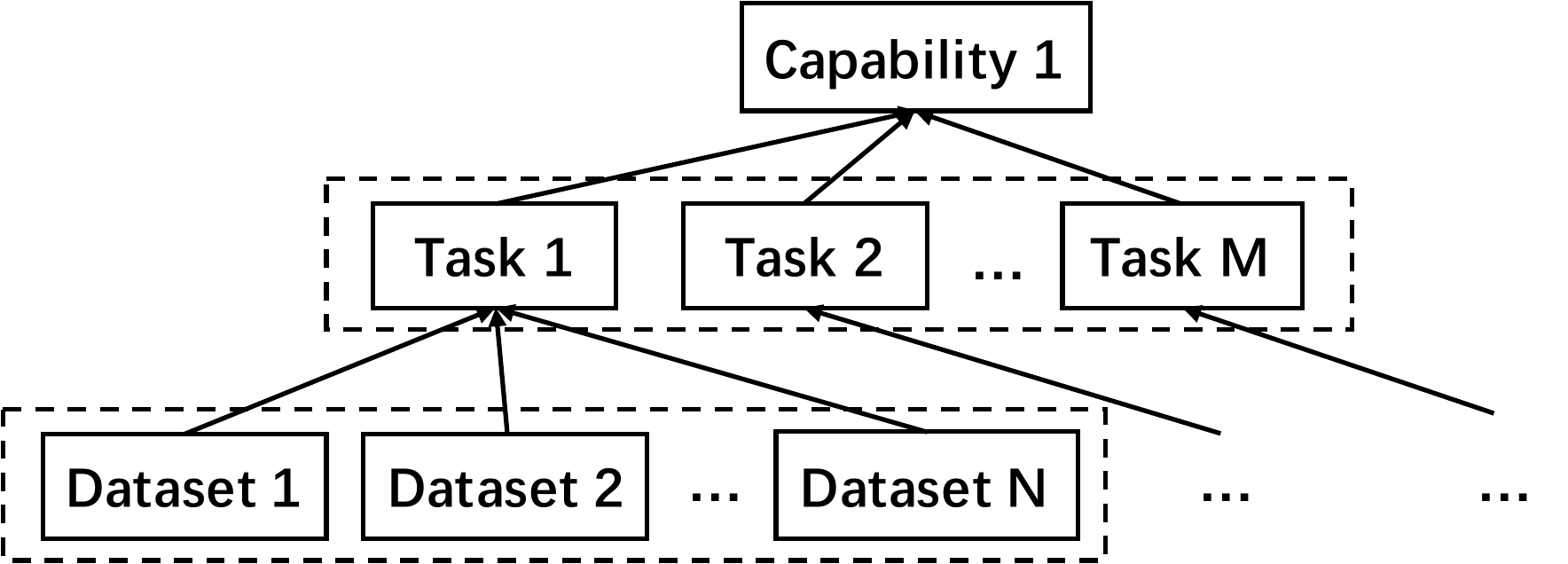}
    \caption{CUGE selects and organizes datasets in a language capability-task-dataset hierarchical framework, based on which multi-level model scores are provided.}
    \label{fig:framework}
\end{figure}

Human language intelligence is general across different language capabilities and tasks. Realizing such general-purpose language intelligence has been a longstanding goal for natural language processing (NLP), where standard evaluation benchmarks play a fundamental and guiding role. To this end, rather than focusing on model performance on specific datasets, researchers have proposed several standard evaluation benchmarks that summarize model performance on a collection of diverse datasets~\cite{wang2018glue,wang2019superglue,xu2020clue,liu2020glge}. Notably, GLUE~\cite{wang2018glue} and SuperGLUE~\cite{wang2019superglue} benchmarks have drawn increasing attention from NLP communities recently and have greatly promoted the development of NLP techniques, especially in the era of pre-trained language models~\cite{devlin2019bert,brown2020language,raffel2020exploring}.

Despite their popularity, there are still important limitations of existing benchmarks. (1) \textit{Flat benchmark framework.} Most existing evaluation benchmarks are dataset-oriented, where commonly used datasets are selected and loosely organized in a flat structure without the guidance of language capabilities to be evaluated. This lack of comprehensiveness and systematicness undermine the reliability of a benchmark as the indicator of general-purpose language intelligence. (2) \textit{Oversimplified scoring strategy.} Most benchmarks summarize the model performance via a simple average of metrics across different datasets, without considering the characteristics of different metrics and datasets. Moreover, since the metrics on all datasets are directly averaged into a single overall value, the performance on language capabilities and tasks cannot be reflected.

To better benchmark general-purpose language intelligence, we propose \textbf{CUGE}, a \textbf{C}hinese Language \textbf{U}nderstanding and \textbf{G}eneration \textbf{E}valuation benchmark with the following features: (1) \textit{Hierarchical benchmark framework.} As shown in Figure~\ref{fig:framework}, CUGE selects and organizes datasets in a language capability-task-dataset hierarchical framework, covering $7$ important language capabilities, $18$ mainstream NLP tasks and $21$ representative datasets. The framework is carefully designed according to the human language examination syllabus and the current NLP research status. With the guidance of the framework, we expect the dataset selection and organization more principled and better reflect general-purpose language evaluation needs. (2) \textit{Multi-level scoring strategy.} Based on the hierarchical framework, in addition to the overall score, CUGE provides model performance evaluation on different levels, including performance on datasets, tasks and language capabilities, more systematically investigating and showing model language intelligence. Moreover, CUGE normalizes the score on each dataset according to the performance of a standard baseline model, largely mitigating the influence of different metrics and datasets.

To facilitate GUGE, we release an online evaluation platform and a public leaderboard. The leaderboard can be easily customized according to model capabilities and tasks in interest to support flexible model judging criteria. Experiments results on representative pre-trained language models show that, although strong performance can be achieved by existing NLP techniques in some language capabilities and tasks, there is still ample room for improvement towards general-purpose language intelligence. The online evaluation platform and leaderboard are publicly available at \url{cuge.baai.ac.cn}.

\section{Design Principles}
The design principles of CUGE are highly correlated to the target of evaluating and innovating general-purpose language intelligence. In this section, we introduce the design principles of the core components of CUGE, including the hierarchical framework, multi-level scoring strategy, dataset evaluation system and the evaluation platform.

\paragraph{Hierarchical Benchmark Framework}
    % 系统：能力-任务-数据集框架
    % 全面：覆盖6个任务40余个数据集
For general-purpose language intelligence evaluation, the benchmark itself needs to be comprehensive and systematic. To this end, different from existing flat dataset-oriented benchmarks built from the bottom up, CUGE is hierarchically constructed in a top-down approach. Specifically, on the top level, according to the human language examination syllabus (i.e., Chinese syllabus for college entrance examination\footnote{\url{www.neea.edu.cn}}) and the current NLP research status, we summarize general-purpose language intelligence into $7$ important language capabilities: (1) language understanding: word-sentence level, (2) language understanding: discourse level, (3) information acquisition and question answering, (4) language generation, (5) conversational interaction, (6) multilingualism, and (7) mathematical reasoning. Then we identify representative NLP tasks for each capability, serving as the task level. Finally, we select representative datasets for each task, resulting in the dataset level. Compared with previous works, the hierarchical benchmark framework of CUGE more systematically organizes existing evaluation resources and more comprehensively reflects general-purpose language evaluation needs.

\begin{table*}[!h]
\centering 
\resizebox{\linewidth}{!}{
\begin{tabular}{lrrrlll}
 \toprule
\textbf{Corpus} & \textbf{$|$Train$|$} & \textbf{$|$Dev$|$} & \textbf{$|$Test$|$} & \textbf{Task} & \textbf{Metrics} & \textbf{Domain} \\

\midrule
\multicolumn{7}{c}{\textbf{Language Understanding: Word-Sentence Level}}\\

PKU-SEG & 40.3k & 10.5k & 9.9k & word segmentation & F1 score & news \\

WordSeg-Weibo & 20.1k & 2.1k & 8.6k & word segmentation & F1 score & social media \\
% PKU & 19.1k & - & 1.9k & Chinese word segmentation & F1 score & news \\

PKU-SEGPOS & 31.7k & 5.2k & 4.8k & word segmentation and POS & F1 score & news \\

CCPM & 21.8k & 2.7k & 2.7k & classical poetry matching & F1 score & Chinese poetry \\

CMeEE & 15.0k & 5.0k & 3.0k & named entity recognition & F1 score & medical \\

FinRE    & 7.5k & 1.5k & 3.7k & relation extraction & F1 score & financial \\

YACLC & 8.0k & 1.0k & 1.0k & grammatical error correction & F1 score & essays \\

% CODT 2.0 & 11.5k    &  2.3k    & 4.8k     & syntactic parsing & LAS  &  mixed-genre      \\ 
% UD-SP  & 3.9K & 0.5k & 0.5k & syntactic parsing & LAS  & Wikipedia \\

% CGED 2017 & 10k & - & 3k & grammatical error diagnosis & accuracy, F1 & education \\
% CGED 2018 & 402k & - & 3.5k & grammatical error diagnosis & accuracy, F1 & education \\

\midrule
\multicolumn{7}{c}{\textbf{Language Understanding: Discourse Level}}\\

SPR &  12.7k   &  1.6k    &  4.8k    &  humor detection & F1 score &  TV show     \\

ClozeT &  0.6k   &  0.3k    &  0.3k    &  story cloze test & accuracy &   literature    \\

C$^3$ &  11.9k   &  3.8k    &  3.8k    &  reading comprehension & accuracy &   mixed-genre    \\ GCRC & 7.0k &  0.8k    &  0.8k    &  reading comprehension & accuracy, F1 score &   mixed-genre    \\
% CMRC 2018 &  10.1k   &  3.2k    &  1.0k    &     reading comprehension & exact match  &   Wikipedia     \\

% Chinese-Literature-RE & 695 & 58 & 84 & relation extraction & F1 score  & literature \\

% THUCNews &  720k    &  16k    &  16k  & text classification & accuracy & news  \\

% JD Full & 3,000k & - & 250k & sentiment analysis & accuracy, F1 &  commercial\\
% ChnSentiCorp & 10k & 1k & 1k & sentiment analysis & accuracy, F1 & mixed-genre\\

% CMNLI & 391k & 12.4k & 13.9k & natural language inference & accuracy & mixed-genre \\
% CSNLI & 545k & 9.3k & 9.2k & natural language inference & accuracy & image caption \\

% Mandarinograd & 415 & 100 & 101 & coreference resolution & accuracy & mixed-genre \\

% ChineseSTS & 10.7k & 1k & 1k & semantic textual similarity & accuracy & mixed-genre \\

% \midrule
% \multicolumn{7}{c}{\textbf{Commonsense Comprehension Capability}}\\

% BabelSememe  & 3.9K & 0.5k & 0.5k & sememe prediction & MAP \& F1  & dictionary \\
% LoCT& 0.5&0.1k&0.3k&commonsense reasoning&accuracy&stories\\

\midrule
\multicolumn{7}{c}{\textbf{Information Acquisition and Question Answering Capability}}\\
WantWords & 78.0k &19.0k &19.0k & reverse dictionary&accuracy@1&mixed-genre\\
KBQA & 24.0k &-& 0.6k &open-domain question answering&EM&mixed-genre\\
Sogou-Log & 8,052k &500k&1.0k&document retrieval&nDCG@k&mixed-genre\\
%OpenCMRC&10k&-&3k&open-domain QA&EM&mixed-genre\\
\midrule
\multicolumn{7}{c}{\textbf{Language Generation Capability}}\\
LCSTS&2,401k&9.0k&0.7k& text summarization & Rouge&news\\
%PersonalDialog&5,418k&10k&10k&personalized dialog&BLEU, Distinct&social media\\
%JDDC&6,490k&5k&5k&task-oriented dialogue&BLEU, Distinct&e-commerce\\
CEPSUM&434k&5.0k&5.0k&text summarization&Rouge&e-commerce\\
E-Reviews&115k&1.0k&3.0k&data-to-text generation&BLEU, Distinct&e-commerce\\
\midrule
\multicolumn{7}{c}{\textbf{Conversational Interaction Capability}}\\
KdConv&62.9k&9.0k&9.1k&knowledge-driven conversation&BLEU, Distinct&film, music, travel\\
\midrule
\multicolumn{7}{c}{\textbf{Multilingual Capability}}\\
WMT20-EnZh&21,000k&4.0k&4.0k&machine translation&BLEU&mixed-genre\\
%XNLI&393k&5k&10k&cross-lingual NLI&accuracy&mixed-genre\\
%XQuAD&88k&11k&1k&cross-lingual QA&F1&mixed-genre\\
%En2ZhSum&1,693k&3k&3k&cross-lingual text summarization&Rouge&mixed-genre\\
NCLS-EnZh&365k&3.0k&3.0k&cross-lingual text summarization&Rouge&mixed-genre\\
\midrule
\multicolumn{7}{c}{\textbf{Mathematical Reasoning Capability}}\\
%Math23k-NC&21k&1k&1k&numerical calculation&accuracy&math word problem\\
Math23k&21.0k&1.0k&1.0k&mathematical computation&accuracy&math word problem\\
%SenPos&20k&0.8k&0.8k&position predition&accuracy&stories\\
%SC&13k&0.5k&0.5k&story completion&BLEU, Distint&stories\\
%OCG&1.5k&0.2k&0.5k&story generation&BLEU, Distinct, Order, Coverage&stories\\
%Story2Moral&3.4k&0.4k&0.4k&moral generation&BLEU, Distinct&Stories\\
%Moral2Story&3.4k&0.4k&0.4k&story generation&BLEU, Distinct, Order, Coverage&stories\\
\bottomrule
\end{tabular}}
\caption{Descriptions and statistics of language capabilities, tasks and datasets.}
\label{tab:tasks}
\end{table*}

% 评测方案
\paragraph{Multi-level Scoring Strategy}
    % 多层次：不是直接平均，逐层汇总任务-能力得分
    % 归一化：与标准模型比较
In addition to a single overall value, we would like to also provide fine-grained, multi-level evaluation of models. Therefore, based on the hierarchical benchmark framework of CUGE, we propose a multi-level scoring strategy for the evaluation results.
% We calculate and aggregate the scores of different levels from the bottom up. 
In addition, it is also desirable to eliminate the influence of different metrics and datasets during score calculation. We thus normalize the scores on datasets based on representative standard baseline models.

% 数据集评价方案
\paragraph{3D Dataset Evaluation System}
The quality of datasets is crucial to a benchmark. However, there is little systematic investigation on this problem in existing benchmarks. To this end, we propose a three-dimensional evaluation system for CUGE's dataset selection and evaluation. Inspired by the assessment of reliability~\cite{cronbach1946case} and validity~\cite{alloway2008evaluating} in educational psychology, we propose \textit{reliability}, \textit{difficulty}, and \textit{validity} as three main dimensions of our dataset evaluation system. We expect the dataset evaluation system will contribute to a more scientific and reliable dataset selection process of CUGE in the future. 

Specifically, (1) \textit{reliability} refers to the consistency and accountability of the scores given by the datasets and the corresponding metrics. 
% We measure the reliability of the datasets by calculating the confidence intervals of the final scores.
(2)~\textit{Difficulty} reflects the hardness and discriminative capability of the datasets. 
% We evaluate the dataset from the difficulty dimension by giving the variation coefficient of different models' performance. 
(3) \textit{Validity} represents the relevance of the evaluation dataset and evaluation goals. The evaluation methods and metrics for dataset quality assessment are still open research problems, and CUGE is actively exploring this direction for building high-quality benchmarks.
% Since quantifying validity can be difficult, we qualitatively select representative tasks and datasets of each language capability, maximally ensuring CUGE datasets are qualified for valid evaluations. %so as to ensure that datasets in CUGE are qualified in the validity dimension.
% Further more, this 3D dataset evaluation system makes CUGE a living benchmark, which means datasets will be evaluated and renewed up to time. It will yield a “moving post” dynamic target for NLP models and can become a moving target for never-ending learning~\cite{mitchell2018never} scenario, rather than a static benchmark that will quickly saturate.

Previous works have shown that evaluation benchmarks need to be continually adjusted according to the current status of NLP research~\cite{wang2018glue,wang2019superglue}. We expect the 3D dataset evaluation system can also serve as a principled criterion for the continual adjustment of datasets in CUGE in the future.

\paragraph{Flexible Online Platform}
    % 可定制化Leaderboard
    % 重诚信
    % 重交互（对模型和数据集的issue讨论区）
To facilitate CUGE, we build an online evaluation platform that features customizable leaderboards and interactive forums. In addition to the leaderboard for each dataset, participants can also customize leaderboards consisting of multiple datasets according to their evaluation objectives. To encourage academic integrity, participants are required to check the honor code before submissions, and interactive forums are provided for submission discussion.

In addition to the evaluation of the existing datasets in CUGE, the platform also supports the release and evaluation of new datasets. Each dataset submitted and released on CUGE will have a dataset-specific leaderboard on the platform. The CUGE benchmark will adjust its dataset collection according to the results of existing and newly released datasets on the platform each year.

\section{Benchmark Framework}
% 具体介绍框架体系
% 具体介绍每个能力、任务、数据集

CUGE selects and organizes datasets in a language capability-task-dataset hierarchical framework, covering $7$ language capabilities, $18$ mainstream NLP tasks and $21$ representative datasets.

% 数据集描述的表格

\subsection{Language Understanding: Word-Sentence Level}

The language understanding: word-sentence level evaluates the model's capability to understand a given text and perform word- and sentence-level syntactic and semantic tasks.

\subsubsection{Word Segmentation}
The task is to identify the sequence of words in a sentence and mark the boundaries between words.

\paragraph{PKU-SEG} The dataset~\citep{emerson-2005-second} is based on the PKU dataset released by The Second International Chinese Word Segmentation Bakeoff in 2005. This dataset is annotated from the news corpus of the People's Daily. We further add the data annotated from the corpus of the People's Daily in January and December 2000. Finally, the data is re-integrated and divided.

\paragraph{WordSeg-Weibo} The dataset~\citep{qiu2016overview} comes from the NLPCC 2016 evaluation task. The dataset is collected from Sina Weibo website\footnote{\url{www.weibo.com}}. Different from the traditional single word segmentation evaluation method, this dataset introduces a new multi-granularity word segmentation evaluation criterion. Besides the training data, we also provide the background data, from which the training and test data are drawn.

\subsubsection{Word Segmentation and POS Tagging}

The task seeks to identify the boundaries between words and assign a pre-defined part-of-speech tag to each word in a given sentence.

\paragraph{PKU-SEGPOS} The dataset is the part-of-speech (POS) tagging dataset collected from the corpus of People's Daily. The corpus from January 2000 and December 1-15, 2020, the corpus from December 16-23, 2000, and the corpus from December 24-31, 2000 compose the training set, the validation set, and the test set, respectively.

\subsubsection{Classical Poetry Matching}

The task is that given a modern description of Chinese classical poetry, the model is supposed to select one from four candidate poems that semantically matches the given description most.

\paragraph{CCPM} The dataset~\citep{li2021CCPM} is a multiple-choice dataset for Chinese classical poetry matching. This dataset comes from the Chinese classical poems and their corresponding modern Chinese translations provided on the website.

\subsubsection{Named Entity Recognition}

The task seeks to locate and classify named entities in unstructured text into pre-defined categories.

\paragraph{CMeEE} The dataset~\citep{hongying2020building} is based on the CHIP2020 evaluation. Given a pre-defined schema, the task is to identify and extract entities from the given sentence and classify them into nine categories: disease, clinical manifestations, drugs, medical equipment, medical procedures, body, medical examinations, microorganisms, and department.

\subsubsection{Entity Relation Extraction}

The task is to extract semantic relations between entity pairs in given sentences.

\paragraph{FinRE} The dataset~\citep{li-etal-2019-FinRE} is a manually labeled financial news relation extraction dataset. Given a sentence and its head and tail entities, the model needs to predict the relation between the head and tail entities. This dataset is annotated from the Sina Finance News corpus, in which the named entity is a commercial company, and the relation contains 44 financial relation categories and an NA category, including special relation categories in the financial and financial fields such as ownership, shareholding, competition, acquisition, transaction, cooperation, and shareholding.

\subsubsection{Grammatical Error Correction}

The task aims to judge the acceptability and correct the grammatical errors of the given text.

\paragraph{YACLC}
The dataset~\cite{wang2021yaclc} is collected from a language learning website~\footnote{\url{lang-8.com}}, where Chinese as foreign language learners share their essays. Given a sentence produced by language learners, models are required to (1) evaluate the grammatical correctness, (2) correct the grammatical errors with minimal edits, and (3) make the sentence more fluent with minimal edits. 

\subsection{Language Understanding: Discourse Level}

The language understanding: discourse level evaluates the model's capability to understand a given text and perform discourse-level syntactic and semantic tasks.

\subsubsection{Humor Detection}

The task is to recognize, classify and generate humor based on computer technology, which has important theoretical and application value.

\paragraph{SPR} The dataset\footnote{\url{cips-cl.org/static/CCL2020/humorcomputation.html}} selects ``I Love My Home" as data source. According to the changes in the scene and plot, the sitcom is divided into several dialogues. In a conversation, there are different characters to communicate, resulting in continuous utterances. Utterances in the same dialogue appear in order and have a contextual relationship. Compared with single-sentence humor, the humor in dialogue may come from the context, rather than the content of the utterance itself. Models are required to judge whether the utterance is humorous based on the context and content, and to identify the punchline in sitcoms.

\subsubsection{Story Cloze Test}

The task is to select the missing sentence from multiple candidates to fill into a story and form a reasonable logical plot.

\paragraph{ClozeT} The dataset comes from children's stories crawled from the Web. When constructing options, crowd-sourced annotators are asked to extract a sentence from the story that can be inferred based on the context and common sense as the correct option, and rewrite it to a sentence contrary to common sense as the wrong option.

\subsubsection{Reading Comprehension}

This task is to answer questions about given unstructured texts.

\paragraph{C$^3$} The dataset~\citep{sun2019investigating} is a multiple-choice Chinese machine reading comprehension dataset, which is collected from test questions for Chinese as a second language learners. Given a Chinese paragraph/dialogue and a question, the dataset provides a number of answer candidates. According to the given content, models are required to choose the correct answers from the candidates.

\paragraph{GCRC} The dataset~\citep{gcrc} is a multiple-choice Chinese machine reading comprehension dataset, which is collected from Chinese college entrance examination. Given an article and a question, models are required to select the correct answers from candidates. The dataset features explainable evaluation for reading comprehension systems. In addition to answering the questions, models are also asked to (1) select the supporting evidence sentences from the article to answer the question, (2) determine the reason for rejecting each wrong answer, and (3) predict the skills required to answer the question.

\subsection{Information Acquisition and Question Answering Capability}
Information acquisition and question answering capability requires retrieving the answer for a query from given non-structured documents or structured knowledge bases.

\subsubsection{Reverse Dictionary}
The reverse dictionary task requires taking the description of a target word as input and outputting the target word.  %together with other words that match the description.
\paragraph{WantWords} The dataset~\cite{zheng2020multi} is constructed based on dictionary definitions, which are collected from Modern Chinese Dictionary (6th Edition). Models are required to select the word described by the text query from the vocabulary.

\subsubsection{Open-domain Question Answering}
Open-domain question answering is the task of answering a question based on a knowledge source.
%\paragraph{OpenCMRC} OpenCMRC~\cite{zhao2020sparta} is a span-extraction based dataset for Chinese machine reading comprehension. The dataset is composed ofof near 20,000 real questions annotated on Wikipedia paragraphs by human experts.
\paragraph{KBQA} The dataset\footnote{\url{tcci.ccf.org.cn/conference/2018/taskdata.php}} is based on the open-domain question answering shared task 7 in NLPCC 2018. Given a natural language question, models are required to produce answers based on the background knowledge base in open domain.

\subsubsection{Document Retrieval} 
The task requires retrieving the relevant document for a given query.
\paragraph{Sogou-Log} The dataset~\cite{luo2017sogout} contains 35 million search sessions with 96k distinct queries collected from a Chinese commercial search engine\footnote{\url{www.Sogou.com}}. %a major Chinese commercial search engine. The sample contains 35 million search sessions with 96k distinct queries.
The query log consists of corresponding queries, displayed documents, user clicks and dwell times. %Each query has an average of 12 documents displayed. 
Each query has 12 documents on average, which tend to be high-quality since the results are collected from a mainstream search engine. 
%As the results come from a commercial search engine, the returned documents tend to be of very high quality. 
The queries in the test sets are sampled from those that appear more than 1k times in all query logs, i.e., head queries. Besides, the dataset uses the performance on the tail queries to evaluate the model robustness.
%Most of the evaluation focuses on the head queries; and the dataset uses tail query performance to evaluate model robustness. 
%The remaining queries are used for model training. The query log contains only document titles and URLs. 

\subsection{Language Generation Capability}
The Language generation capability requires generating readable natural language texts conditioned on given inputs.
\subsubsection{Text Summarization}
The task requires generating a short text to include the important information of a given long text.
\paragraph{LCSTS} The dataset~\cite{hu2015lcsts} is a text summarization dataset collected from Sina Weibo. This corpus contains more than 2M Chinese texts paired with short summaries, which are typically written by the corresponding authors. LCSTS also provides 10.7k summaries that are manually annotated with relevance scores to the corresponding texts. We only take those examples whose relevance scores are larger than 2 as the test set.

\paragraph{CEPSUM} The dataset~\cite{yuan-etal-2020-faithfulness} is a collection of product summaries on a mainstream Chinese e-commerce platform JD. The products come from two categories including home appliances and bags. The summaries are generated by thousands of experts, and the auditing groups of the e-commerce platform verify the quality. 

\subsubsection{Data-to-Text Generation}
The task requires generating natural language texts from structured data.
\paragraph{E-Reviews} The dataset~\cite{shao-etal-2019-long} is collected from a Chinese e-commerce platform. Given a table that contains a set of attribute-value pairs to describe a commodity, models are asked to produce natural language advertising texts. 
\subsection{Conversational Interaction Capability}
The task requires generating fluent and reasonable responses to users' posts.
\subsubsection{Knowledge-driven Conversation Generation}
\paragraph{KdConv} The dataset~\cite{zhou2020kdconv} is collected for research on knowledge-driven multi-turn conversation. %, which is suitable for modeling knowledge interactions in multi-turn human-like dialogues, including knowledge planning, knowledge grounding, knowledge adaptations, etc. 
KdConv consists of 4.5k dialogues that cover three domains including film, music, and travel, where each domain contains 1.5k dialogues. Therefore, the dataset can be used to explore knowledge transfer among these domains. Besides, these dialogues cover diversified topics and are manually annotated with related knowledge facts. %ranged from one to four, without any pre-defined goals or constraints, which are closer to real human-human conversations than other datasets. 

\iffalse
\subsubsection{Personalized Dialogue}
\paragraph{PersonalDialog} PersonalDialog~\cite{zheng2020personalized} involves a large number of speakers with a wide variety of personality traits. The data in PersonalDialog are collected from \url{weibo.com}, one of the largest Chinese social media. Each speaker presented in PersonalDialog has five personality traits: ``Gender'', ``Age'', ``Location'', ``Interest Tags'', and ``Self Description''. Specifically, ``Gender'' is a binary-valued trait, i.e., the gender of a speaker can be either ``Male'' or ``Female''; ``Age'' is represented by an integer ranging from 8 to 48. Our observation indicates that Age values out of this range are very likely to be ``fake'', i.e., some Weibo users prefer not to reveal their true ages by providing unreasonable birthdays.

\subsubsection{Task-Oriented Dialogue}
\paragraph{JDDC} JDDC~\cite{chen-etal-2020-jddc} is a large-scale real scenario Chinese E-commerce conversation corpus collected from JD, with
more than 1 million multi-turn dialogues, 20 million utterances, and 150 million words. The dataset reflects several characteristics of human-human conversations, e.g., goal-driven, and long-term dependency among the context. It also covers various dialogue types including task-oriented, chitchat and question-answering. We formulate the task as generating an answer for a given question paired with the previous dialogue with at most four turns between the user and the service.

\fi

\subsection{Multilingual Capability}
Multilingual capability requires handling inputs and outputs in multiple languages. 
\subsubsection{Machine Translation}
The task requires translating a natural language text to another specified language and maintaining the semantics.
\paragraph{WMT20-EnZh} The dataset~\cite{barrault-EtAl:2020:WMT1} is based on the shared task of the workshop on machine translation in 2020, where the corpus is collected from news websites. We focus on machine translation between Chinese and English. The test set for English to Chinese is produced by translating at the paragraph level. 

\iffalse
\subsubsection{Cross-Lingual Natural Language Inference} 
\paragraph{XNLI} XNLI~\cite{conneau2018xnli} is proposed for evaluating cross-lingual natural language inference, which consists of 7500 human-annotated development and test examples in NLI three-way classification format in English, French, Spanish, German, Greek, Bulgarian, Russian, Turkish, Arabic, Vietnamese, Thai, Chinese, Hindi, Swahili and Urdu, making a total of 112,500 annotated pairs. These languages span several language families, and with the inclusion of Swahili and Urdu, include two lower-resource languages as well.
\subsubsection{Cross-Lingual Question Answering}
\paragraph{XQuAD} XQuAD~\cite{artetxe2019cross} consists of a subset of 240 paragraphs and 1190 question-answer pairs from the development set of SQuAD v1.18~\cite{rajpurkar-etal-2016-squad} together with their translations into ten languages: Spanish, German, Greek, Russian, Turkish, Arabic, Vietnamese, Thai, Chinese, and Hindi. Both the context paragraphs and the questions are translated by professional human translators. %In order to facilitate easy annotations of answer spans, we choose the most frequent answer for each question and mark its beginning and end in the context paragraph using special symbols, instructing translators to keep these symbols in the relevant positions in their translations.
\fi
\subsubsection{Cross-lingual Text Summarization}
Given a document from the source language, the task requires generating a short text that summarizes the key information in the target language.

\paragraph{NCLS-EnZh} The dataset~\cite{zhu-etal-2019-ncls} is constructed based on round-trip translation, where the corpus comes from CNN, DailyMail and Sina Weibo websites. The reference texts in the test set are further corrected by multiple human annotators. Here we also focus on the English-to-Chinese summarization. %~(called NZh2EnSum) and vice versa~(called En2ZhSum). 

\newcommand{\tc}[1]{\multicolumn{1}{c}{#1}} %居中对齐
\newcommand{\tl}[1]{\multicolumn{1}{l}{#1}} %左对齐
\newcommand{\tr}[1]{\multicolumn{1}{r}{#1}} %右对齐
\begin{table*}[t]
\resizebox{\linewidth}{!}{%
\small
\centering
\begin{tabular}{lcccccccc}
\toprule
      & NLU-WSL  & NLU-DL   & IA\&QA & NLG & CI         & ML & MR &  CUGE Index\\ \midrule
    %   & Acc & Acc  & BLEU    & Acc     & MRR  & Rouge-L   \\
% mT5-XXL   &    90.62     & 86.36    &     23.98       & 59.27   & 35.37    &    34.80  & 55.81  \\
mT5-Small   &   87.70 (100)     & 41.50 (100)  &  29.20 (100) & 33.10 (100) & \tr{8.76 (100)} & \tr{9.10 (100)} & 18.40 (100) & 100  \\
mT5-Large 
& 89.90 (103)
& 56.30 (136)
& 31.65 (108)
& 34.40 (104)
& \tr{9.76 (111)}
& 11.10 (122)
& 34.30 (186)
& 117 \\
mT5-XXL
& 90.60 (103)
& 86.40 (208)
& 35.90 (123)
& 34.80 (105)
& 12.68 (145)
& 24.00 (264)
& 61.60 (335)
& 152 \\
CPM-2 & 91.60 (104)
& 86.10 (207)
& 35.90 (123)
& 35.90 (108)
& 13.12 (150)
& 26.20 (288)
& 69.40 (377)
&  157 \\
\bottomrule
\end{tabular}}
\caption{Performance of representative pre-trained language models on lite version of CUGE (\%). In addition to the raw performance, we also report the performance normalized by mT5-small (in parenthese), based on which the overall CUGE Index is calculated. NLU-WSL: language understanding: word-sentence level, NLU-DL: language understanding: discourse level, IA\&QA: information acquisition and question answering, NLG: language generation, CI: conversational interaction, ML: multilingualism, and MR: mathematical reasoning.}
\label{tab:main results}
\end{table*}

\subsection{Mathematical Reasoning Capability}
The capability requires inferring the answer to a math word problem. 
\subsubsection{Mathematical Computation} 
\paragraph{Math23k} The dataset~\cite{wang2017deep} is a collection of math word problems from online education websites. The problems can be solved based on knowledge of linear algebra with only one variable. In addition to the answer, each problem in the dataset is also annotated with an equation for solving the problem.

\section{Model Scoring Strategy}
% 评测方案：多层次、归一化
In terms of model scoring, we adopt a multi-level scoring and model-based normalization strategy.

\paragraph{Multi-level Scoring} Based on the hierarchical benchmark framework, we aggregate model scores from the bottom up. Based on the scores on the dataset level (e.g., F1 score and accuracy), the score of the corresponding task can be obtained from the normalized average of dataset scores. Similarly, the capability scores are the average of task scores, and the overall CUGE score is the average of capability scores. Through this strategy, we put forward a multi-level, fine-grained evaluation of the whole capability-task-dataset framework.

\paragraph{Score Normalization} Existing NLP benchmarks typically calculate overall scores using the average of model scores on datasets. However, this procedure neglects characteristics of different datasets and metrics. For instance, the numerical results of BLEU~\cite{papineni2002bleu} are typically small as compared with F1 scores. As a result, with score averaging, the overall score can be dominated by metrics with larger scale. 

To address the issue, we normalize the scores based on scores of representative standard baseline models, i.e., mT5-Small~\cite{xue2020mt5}, so as to eliminate the influence of disturbing factors such as different metrics. Specifically, the normalized score on dataset is given by $p/b$, where $p$ and $b$ are dataset performance of model under evaluation and the standard baseline model respectively. Normalization on standard model performance essentially gives a ratio to different metrics and datasets, therefore making the overall score more reasonable.

\begin{table}[t]
\scriptsize
\centering
\resizebox{\linewidth}{!}{
\begin{tabular}{lll}
\toprule
Capability & Task & Dataset \\
\midrule
NLU-WSL & Classical Poetry Matching & CCPM \\
NLU-DL & Reading Comprehension & C$^3$ \\
IA\&QA & Document Retrieval & Sogou-Log \\
NLG & Text Summarization & LCSTS \\
CI & Conversation Generation & KdConv \\
ML & Machine Translation & WMT20-EnZh \\
MR & Mathematical Computation & Math23K \\
\bottomrule
\end{tabular}}
\caption{Lite version of CUGE. Each capability is instantiated with the most representative task and dataset. NLU-WSL: language understanding: word-sentence level, NLU-DL: language understanding: discourse level, IA\&QA: information acquisition and question answering, NLG: language generation, CI: conversational interaction, ML: multilingualism, and MR: mathematical reasoning.}
\label{tab:lite}
\end{table}

\section{Using CUGE}
To facilitate CUGE, we build a public online evaluation platform. Participants can view the benchmark framework and leaderboard, download datasets, and participate in the evaluation by submitting prediction files. Specifically, CUGE platform enjoys the following notable features:

\begin{figure*}[t]
    \centering
    \includegraphics[width=\textwidth]{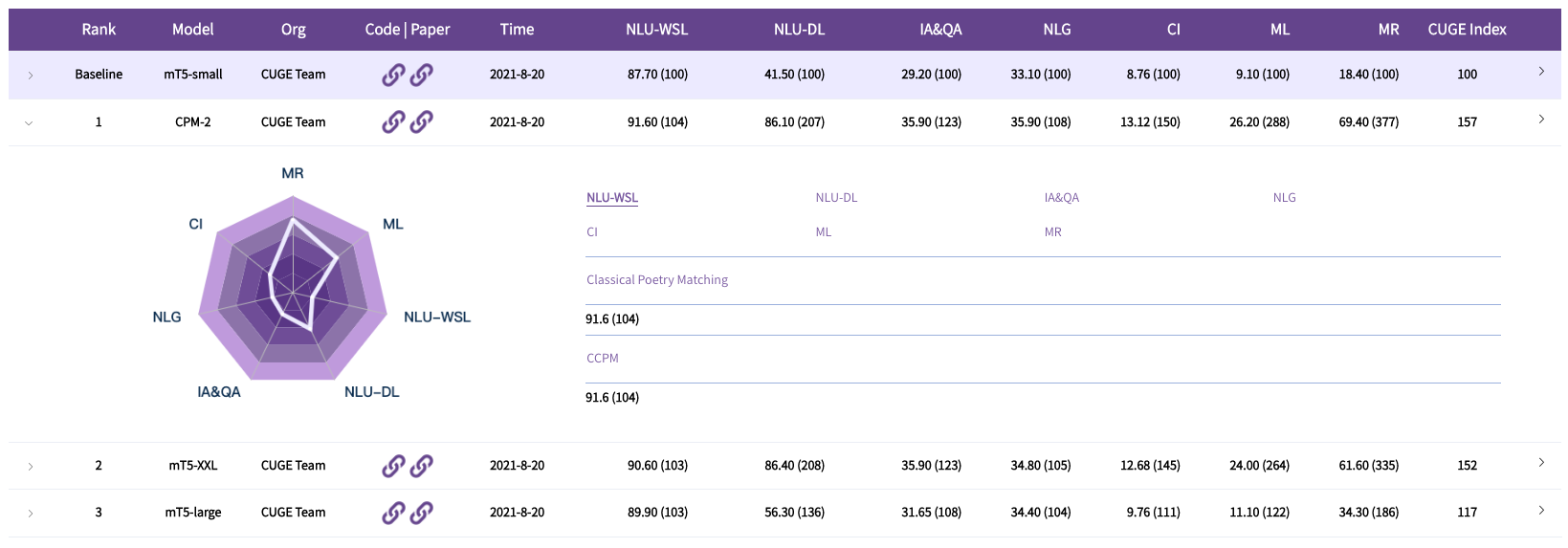}
    \caption{Lite leaderboard of CUGE platform.}
    \label{fig:platform}
\end{figure*}

\begin{figure*}[t]
        \centering
        \subfloat[mT5-Large]{\includegraphics[width=0.33\textwidth]{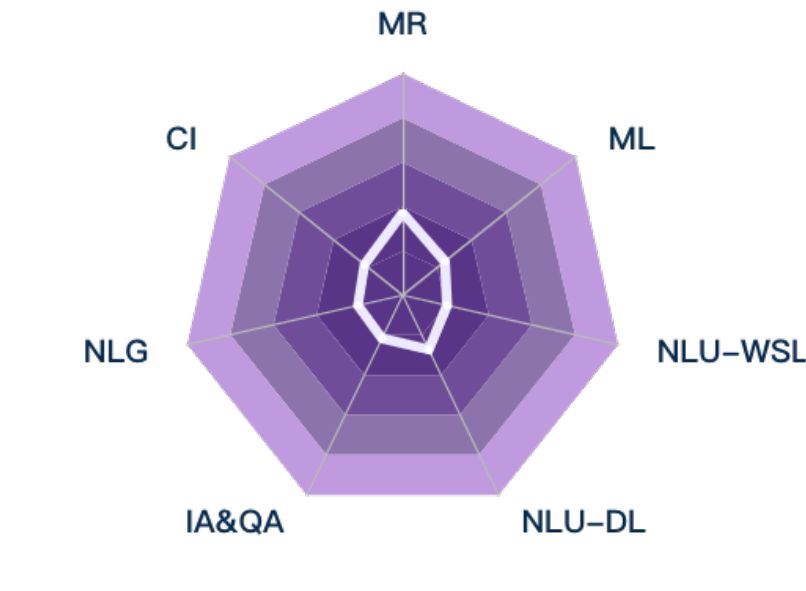}\label{fig:case1}}
        \subfloat[mT5-XXL]{\includegraphics[width=0.33\textwidth]{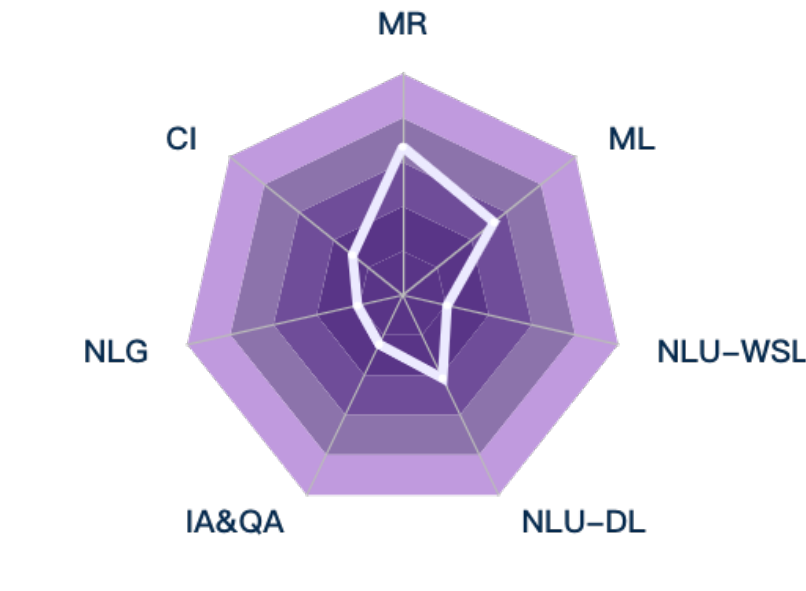}\label{fig:case2}}
        \subfloat[CPM-2]{\includegraphics[width=0.33\textwidth]{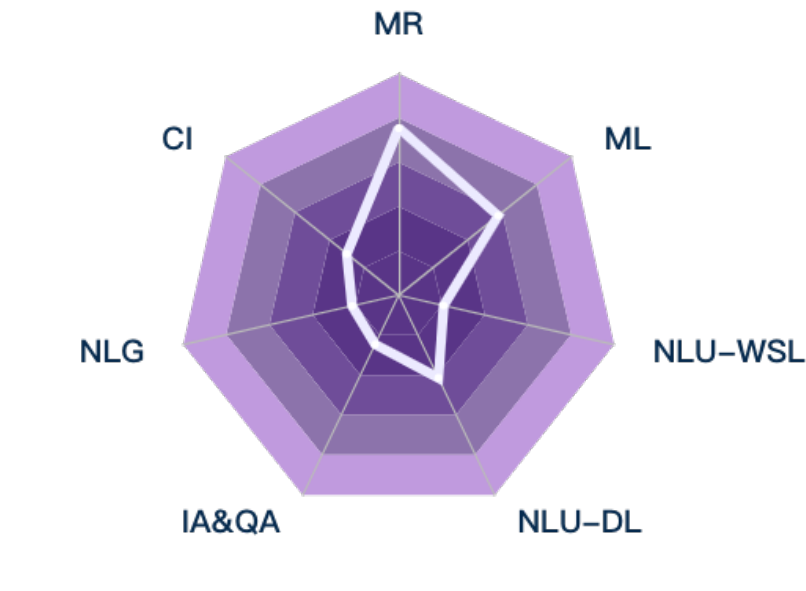}\label{fig:case3}}
        \caption{Capability performance visualization for representative pre-trained language models on CUGE.}
        \label{fig:radar}
\end{figure*}

\paragraph{Customizable Leaderboard}
CUGE characterizes datasets with multi-dimensional labels, such as language capability and task. Users can customize the leaderboard by selecting the labels, supporting flexible model judging criteria. CUGE also recommends a standard lite leaderboard, as shown in Table~\ref{tab:lite}. Specifically, for each language capability, we first select the most representative task and the most representative dataset of this task, and then combine the datasets to build the lite leaderboard, which enables convenient and rapid evaluation on GUGE platform, as shown in Figure~\ref{fig:platform}.

\paragraph{Academic Integrity Encouragement}
As a standard evaluation benchmark, CUGE highly values academic integrity of evaluation participants. Therefore, participants are required to check the honor code before submissions, which forbids usage of human labels on the CUGE test set in any form. Participants are also encouraged to release technique reports and codes on the leaderboard.

\paragraph{Interactiveness}
In the future, CUGE will provide official forums for users to encourage discussion on: (1) Submissions on leaderboard, which helps better understanding of submissions and supports supervision for academic integrity; (2)~Datasets in CUGE, which helps the adjustment of CUGE datasets.

% 介绍网站
% 可定制化Leaderboard
% 提交规范
% 重诚信
% 重交互（对模型和数据集的issue讨论区）

\section{Experiments}

In this section, we present experimental results of representative baseline models on CUGE.

\subsection{Experimental Setup}
In our experiments, we select mT5-Small~\cite{xue2020mt5}, a representative pre-trained language model with $300$M parameters as our standard baseline model to normalize the performance of models under evaluation. mT5 adopts an encoder-decoder structure, and is capable of both language understanding and generation, making it a good standard baseline model for CUGE. Based on the normalization of mT5-Small, we evaluate mT5-Large (1.2B), mT5-XXL (13B), and CPM-2 (11B)~\cite{zhang2021cpm} on the lite version of CUGE.  

\subsection{Results}
We report the experimental results in Table~\ref{tab:main results} and visualize the normalized capability performance in Figure~\ref{fig:radar}, from which we have the following observations: (1) mT5-XXL significantly outperforms mT5-Large and mT5-small, showing that increased number of parameters can lead to better performance in different language capabilities. (2)~With smaller model size, CPM-2 outperforms mT5-XXL, which shows that improved pre-training procedure can produce stronger language capabilities of pre-trained language models. (3)~However, the performance improvement over different language capabilities is highly imbalanced. For example, the improvement of language generation capability is substantially smaller than multilingual capability. Such overall evaluation and investigation of language capabilities cannot be achieved in existing benchmarks. The results show the necessity of constructing a comprehensive and systematic evaluation benchmark to better evaluate and guide general-purpose language intelligence research.

% \begin{figure}[t]
%     \centering
%     \includegraphics[width=0.9\columnwidth]{acl-ijcnlp2021-templates/figures/Radar.pdf}
%     \caption{Language capability scores of mT5-XXL on lite version of CUGE. }
%     \label{fig:radar}
% \end{figure}

\section{Conclusion and Future Work}
In this work, we present CUGE, a Chinese language understanding and generation evaluation benchmark. CUGE features capability-task-dataset hierarchical framework and multi-level scoring strategy. To facilitate CUGE, we build a public online evaluation platform that supports customizable leaderboards. Experimental results on representative pre-trained language models indicate ample room for improvement towards general-purpose language intelligence. 

Note that CUGE and its platform are still in progress. The current version of CUGE mainly adopts existing datasets for model evaluation, and the platform only implements part of the ultimate design. In the future, we plan to (1) continually build high-quality datasets tailored for the evaluation objective of CUGE, (2) conduct more detailed dataset quality evaluation, and (3) fully implement the design of CUGE platform.

\section{Acknowledgements}
We would like to thank Beijing Academy of Artificial Intelligence (BAAI) for providing support.

\bibliographystyle{acl_natbib}
\bibliography{acl2021}

%\appendix

\appendix

\section{Contributions}

\noindent \textbf{Yuan Yao, Qingxiu Dong, Jian Guan and Boxi Cao} built the benchmark and wrote the report. Yuan Yao wrote Section 1, 4, 5, 6 and 7; Qingxiu Dong wrote Section 2; Jian Guan wrote Section 3.3-3.7; Boxi Cao wrote Section 3.1-3.2. Yuan Yao and Zhiyuan Liu proofread the report.

\vbox{}

\noindent \textbf{Yuan Yao, Chengqiang Xu, Huadong Wang and Guoyang Zeng} led the platform design and implementation.

\vbox{}

\noindent \textbf{Yuan Yao, Qingxiu Dong, Jian Guan, Boxi Cao, Xiaozhi Wang, Fanchao Qi, Junwei Bao, Jinran Nie, Zheni Zeng, Xuancheng Huang, Kun Zhou, Wenhao Li, Shuhuai Ren, Jinliang Lu and Zile Zhou} reviewed, selected and organized the datasets.

\vbox{}

\noindent \textbf{Zhengyan Zhang, Chaojun Xiao and Yuxian Gu} conducted the experiments.

\vbox{}

\noindent \textbf{Zhiyuan Liu, Xianpei Han, Erhong Yang, Zhifang Sui and Maosong Sun} designed and led the research.

\vbox{}

\noindent \textbf{Jiajun Zhang, Juanzi Li, Minlie Huang, Rui Yan, Xiaodong He, Xiaojun Wan, Xin Zhao, Xu Sun and Yang Liu} participated in the discussions of the research and provided valuable suggestions.

\vbox{}

\end{document}